\documentclass{article}
\usepackage[preprint]{neurips_2024}
\pdfoutput=1
\usepackage[utf8]{inputenc} 
\usepackage[T1]{fontenc} 
\usepackage{hyperref} 
\usepackage{authblk}
\usepackage{url}  
\usepackage{booktabs} 
\usepackage{natbib}
\setcitestyle{numbers,square}
\usepackage{amsfonts}       
\usepackage{nicefrac}       
\usepackage{microtype}      
\usepackage{xcolor}
\usepackage{amsmath}
\usepackage{footnote}
\usepackage{amssymb}
\usepackage{algorithm}
\usepackage{algorithmic}
\usepackage{textcomp}
\usepackage{subfigure}
\usepackage{booktabs}
\usepackage{multirow}
\usepackage{tipa}
\usepackage{url}
\usepackage{float}
\usepackage{graphicx} %
\usepackage{soul}
\usepackage{color, xcolor}

\makeatletter 
\renewcommand{\@thesubfigure}{\hskip\subfiglabelskip}

\title{Cross-domain Few-shot In-context Learning for \\ Enhancing Traffic Sign Recognition}

\author{
  \textbf{Yaozong Gan}$^{\dagger}$ \qquad \textbf{Guang Li}$^{\ddagger}$ \qquad \textbf{Ren Togo}$^{\dagger\dagger}$ \\ 
  \vspace{0.3cm}
  \textbf{Keisuke Maeda}$^{\dagger\dagger\dagger}$ \qquad \textbf{Takahiro Ogawa}$^{\dagger\dagger}$ \qquad \textbf{Miki Haseyama}$^{\dagger\dagger}$
}

\affil{
  $^{\dagger}$Graduate School of Information Science and Technology, Hokkaido University, Japan \\
  $^{\ddagger}$Education and Research Center for Mathematical and Data Science, Hokkaido University, Japan \\
  $^{\dagger\dagger}$Faculty of Information Science and Technology, Hokkaido University, Japan \\
  $^{\dagger\dagger\dagger}$Data-Driven Interdisciplinary Research Emergence Department, Hokkaido University, Japan \\
    \vspace{0.3cm}
\texttt{\{gan, guang, togo, maeda, ogawa, mhaseyama\}@lmd.ist.hokudai.ac.jp}
}

\begin{document}

\maketitle

\begin{abstract}
Recent multimodal large language models (MLLM) such as GPT-4o and GPT-4v have shown great potential in autonomous driving.
In this paper, we propose a cross-domain few-shot in-context learning method based on the MLLM for enhancing traffic sign recognition (TSR). We first construct a traffic sign detection network based on Vision Transformer Adapter and an extraction module to extract traffic signs from the original road images. To reduce the dependence on training data and improve the performance stability of cross-country TSR, we introduce a cross-domain few-shot in-context learning method based on the MLLM. To enhance MLLM's fine-grained recognition ability of traffic signs, the proposed method generates corresponding description texts using template traffic signs. These description texts contain key information about the shape, color, and composition of traffic signs, which can stimulate the ability of MLLM to perceive fine-grained traffic sign categories. By using the description texts, our method reduces the cross-domain differences between template and real traffic signs. Our approach requires only simple and uniform textual indications, without the need for large-scale traffic sign images and labels. We perform comprehensive evaluations on the German traffic sign recognition benchmark dataset, the Belgium traffic sign dataset, and two real-world datasets taken from Japan. The experimental results show that our method significantly enhances the TSR performance.
\end{abstract}

\section{Introduction}

\par Traffic safety is an important issue in the real world. According to the latest statistics from the World Health Organization, about 1.19 million people die in road traffic accidents every year. In addition, road traffic injuries are the leading cause of death for children and young people between the ages of 5 and 29~\footnote{https://www.who.int/news-room/fact-sheets/detail/road-traffic-injuries}.
Moreover, road traffic accidents cause serious economic losses and create a burden on society~\cite{hammoudi2022road}. Therefore, it is urgent to reduce the occurrence of road traffic accidents.
\par Understanding and recognizing traffic signs is vital for traffic safety. This task is challenging, particularly in complex weather and road conditions~\cite{seraj2021implications}. Advanced driver assistance systems (ADAS) use traffic sign data to evaluate driving conditions and notify drivers of inconsistencies, which is crucial for enhancing vehicle safety in emergency situations~\cite{romdhane2016improved}. Furthermore, TSR can aid global positioning system (GPS) service providers in updating their mapping databases and help with intelligent traffic management~\cite{feng2024novel}. Therefore, it is important to study effective TSR techniques.

\par Early studies on traffic sign recognition (TSR) focus on the use of hand-crafted features and traditional image processing methods~\cite{dilek2023computer}. Hand-crafted features are obtained through a variety of algorithms using information in the images. Examples include TSR methods based on the histogram of oriented gradients (HOG)~\cite{yucong2021traffic} and scale-invariant feature transform (SIFT)~\cite{lowe2004distinctive}. Newer methods are based on convolutional neural networks (CNNs)~\cite{saadna2017overview}, which are trained on existing datasets to learn the features of traffic signs, and have achieved high recognition accuracy on datasets from specific countries. 
However, accurate recognition using these CNN-based methods requires careful training on country-specific datasets. When there is a lack of sufficient training data, the effectiveness of these methods remains unknown. Besides, the Vienna Convention on Road Traffic~\cite{economic1968convention} specifies more than 300 different traffic sign categories, and 83 countries have signed the treaty, but there are still some visual differences between the traffic sign images of each country. Therefore, these methods require training data for transfer learning, especially when traffic signs vary between countries.
Second, solutions for TSR through unsupervised learning and feature matching have also been proposed to address the problems mentioned above~\cite{ren2009general,gan2023zero,1520577215790999680}. These methods perform TSR without training data, which solves the problem of cross-country traffic sign inapplicability that exists with supervised methods. However, the recognition accuracy of these methods needs to be further improved.

\par Recently, multimodal large language models (MLLMs) have made significant progress~\cite{zhao2023survey}. In particular, since the release of GPT-4, MLLMs have attracted increasing interest from the research community. Many works have been devoted to constructing multimodal GPT-4 based on open-source models~\cite{zhang2023llama}. In particular, GPT-4o and GPT-4 with vision (GPT-4v) has powerful multi-modal perception and reasoning capabilities. In addition, there are also some preliminary studies on few-shot in-context learning methods based on GPT-4v to stimulate the potential to recognize complex images~\cite{yang2023dawn}. However, despite GPT-4v's unprecedented visual language understanding capabilities, its fine-grained recognition capabilities for small objects such as traffic signs remain to be explored. It is reported that GPT-4v has difficulty generating accurate coordinate series of small objects such as traffic signals in images~\cite{you2023ferret}, which reflects its difficulty in identifying small objects in complex road images. In general tasks, the original images are usually directly input into MLLM for recognition. However, due to the complex composition of road images and the need for fine-grained recognition of different types of traffic signs in TSR, it is necessary to perform detailed research to explore suitable TSR methods based on MLLM. Furthermore, to obtain better performance, the prompt of MLLM must be carefully designed. Therefore, how to use the characteristics of traffic signs to enhance the TSR capability of MLLM is worthy of study.
 
\par In this paper, we introduce a novel cross-domain few-shot in-context learning method to enhance TSR performance. We propose a traffic sign detection (TSD) network for extracting traffic signs from real-world road images. The proposed TSD network includes the development of a dedicated traffic sign detection module based on Vision Transformer Adapter (ViT-Adapter)~\cite{chen2023vision} and a traffic sign extraction module. We propose a novel strategy based on few-shot in-context learning for generating description texts. The description texts generated from template traffic signs contain key information such as color, shape, and composition, which can optimize the ability of MLLM to recognize traffic signs. The proposed few-shot in-context learning strategy requires only traffic sign template images, which can be easily obtained in the national common traffic sign template database. Since there are cross-domain differences between template traffic signs and actual traffic signs, our approach reduces the cross-domain differences by utilizing MLLM to generate description texts. It is worth mentioning that our cross-domain few-shot approach relies on simple and standardized textual indications, which performs effective TSR across countries without the need for large-scale training images. Our contributions are concluded as follows.
\begin{itemize}
    \item We propose a novel few-shot in-context learning method based on the MLLM for enhancing TSR.
    \item We introduce a cross-domain method to reduce the cross-domain difference between traffic sign templates and real traffic signs by using MLLM to generate text descriptions.
    \item We realize promising TSR results on the two benchmark datasets and two real-world datasets.
\end{itemize}
\section{Relate Works}
\subsection{Traffic Sign Recognition}
\par TSR has been extensively studied, yielding various approaches to address this task. TSR is mainly divided into two steps: traffic sign detection (TSD) and traffic sign classification (TSC). TSD localizes traffic signs from road images and TSC classifies and recognizes the detected traffic signs. In this subsection, we provide an overview of the related works on traditional methods and deep learning-based TSR methods.

The first research on TSR was proposed in the 1980s when researchers attempted to create early TSR systems~\cite{akatsuka1987road}. These systems performed visual inspections, recognized traffic signs, and transmitted the information to drivers, alerting them to specific signs. Subsequently, TSR methods based on hand-crafted features and machine learning algorithms were proposed~\cite{mathias2013traffic}. Hand-crafted features were used to extract information from traffic signs, and machine learning algorithms recognized the extracted features. Kus et al.~\cite{kus2008traffic} proposed a traffic sign detection and recognition technique that enhanced SIFT~\cite{lowe2004distinctive} by incorporating features related to the color of local regions. Hu et al.~\cite{hua2010traffic} proposed a TSR method based on SIFT and support vector machine (SVM)~\cite{chen2005tutorial}. SIFT detects and characterizes key points of traffic signs, and SVM classifies the signs. Traditional TSR methods rely heavily on hand-crafted features, which are sensitive to changes in lighting conditions, occlusion, and complex backgrounds~\cite{kerim2021recognition}. Besides, these methods are difficult to adapt to diverse traffic sign datasets and real-world scenarios, as hand-crafted features may not generalize well~\cite{li2022finely}. Despite these limitations, traditional methods remained the basis of early TSR research.
The emergence of deep learning has inspired the development of TSR. Compared with traditional hand-crafted feature-based methods, deep learning-based methods using CNNs can learn low-level visual features and high-level semantic information from traffic sign images. Luo et al.\cite{luo2017traffic} proposed a data-driven integrated TSR system, achieving TSR by refining and classifying regions of interest using a multi-task CNN. Zhu et al.~\cite{zhu2016traffic} proposed an end-to-end CNN-based method for TSR. Besides, Zheng et al.~\cite{zheng2022evaluation} performed a detailed evaluation of TSR using vision transformer (ViT)~\cite{dosovitskiy2020image}. Although these deep learning-based TSR methods perform well, they usually require careful training on country-specific datasets. Several approaches have been introduced to reduce the dependence on training data. Supriyanto et al.~\cite{supriyanto2016unsupervised} proposed an unsupervised approach based on the bag-of-visual-word model, which does not require label data and a training process for TSR. Gan et al.~\cite{gan2023zero} proposed a zero-shot method based on midlevel features of CNNs. This method performs TSR by calculating the similarity between the target and template traffic signs. These methods solved the problem of cross-country traffic sign inapplicability that exists with supervised methods. However, the recognition accuracy of these methods needs to be further improved.

\subsection{ Multimodal Large Language Models}
Recently, MLLMs have attracted attention in both academia and industry~\cite{chang2023survey}. As demonstrated by the existing work~\cite{bubeck2023sparks}, MLLMs can solve a wide variety of tasks, which contrasts with previous models that were limited to solving specific tasks. MLLMs are increasingly being used due to their excellent performance in handling different applications such as general natural language tasks and domain-specific tasks. In addition, the proposal of multimodal large language models such as GPT-4 further extends the functionality of language models by seamlessly integrating visual information as part of the input. This integration of visual data enables the models to effectively understand and generate responses that contain both textual and visual cues, thus enabling contextually richer conversations in multimodal environments. In recent months, MLLMs have also received a lot of attention in the field of intelligent transportation, such as autonomous driving and mapping systems~\cite{ cui2024drive}. However, to the best of our knowledge, there is no related research on MLLMs in the field of TSR. Therefore, we implement a prior study of MLLMs in TSR to explore an effective method in this paper.

\section{Methodology}
\begin{figure*}[t]{
    \centering
    \includegraphics[width=0.9\textwidth]{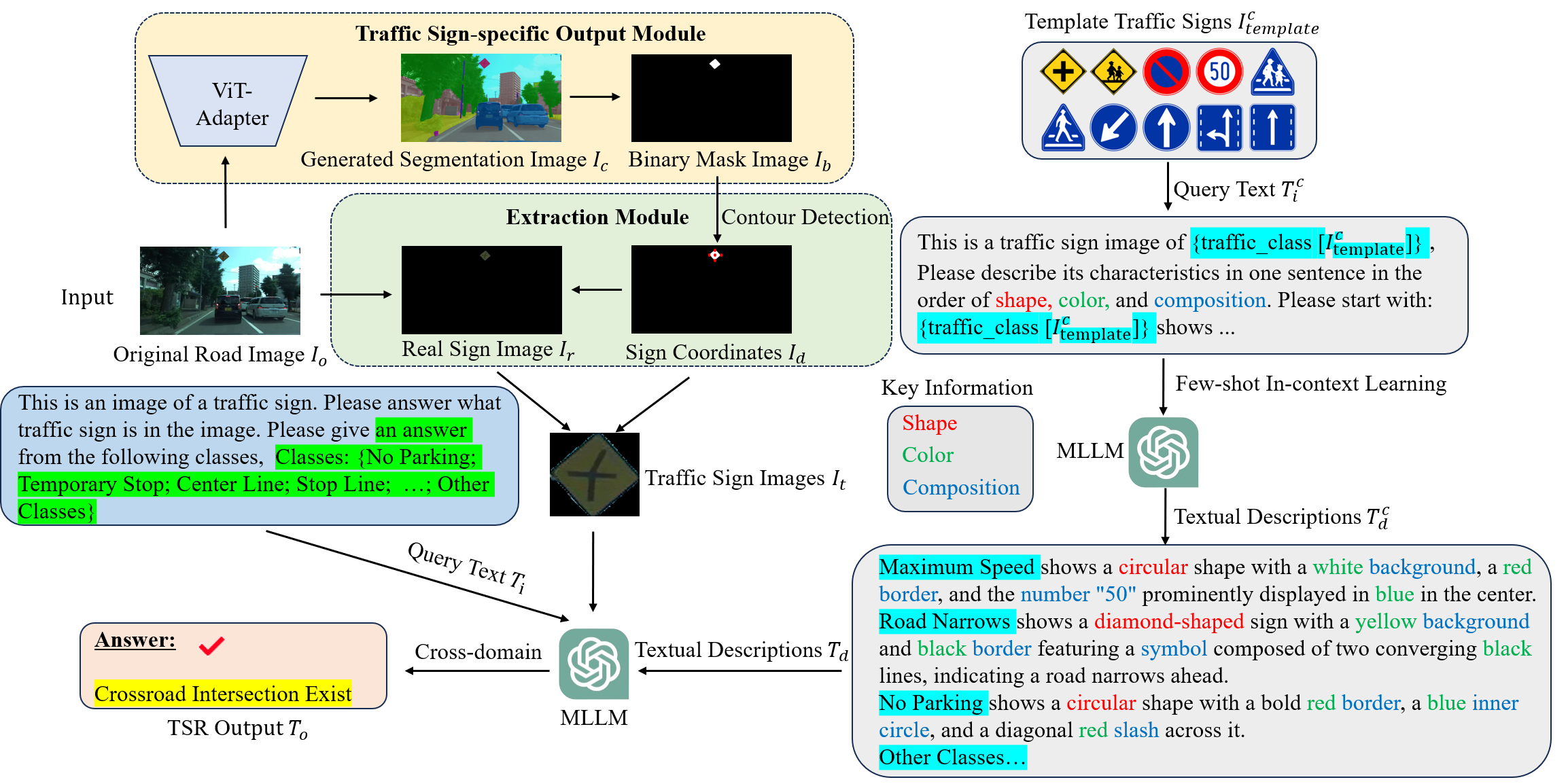}
    \caption{Overview of the proposed cross-domain few-shot in-context learning TSR method. We first perform traffic sign detection and extraction based on the proposed TSD network. Then we use template traffic signs to generate the description texts based on MLLM. The generated description texts contain key information about the shape, color, and composition of traffic signs, thus improving MLLM's reasoning ability for traffic signs.} \label{fig1}}
\end{figure*}

In this section, we introduce the proposed method. As shown in Figure~\ref{fig1}, our method performs TSD from original road images and fine-grained TSC. By using template traffic signs in the national standard template database to construct a cross-domain few-shot in-context learning method with MLLM, the proposed method eliminates the dependence on large-scale training data with labels.

\subsection{Traffic Sign Detection}\label{sub3.1}
\par TSD is the first phase of the proposed method, aimed at the rapid localization and detection of traffic signs in original urban road images. Our approach is inspired by ViT-Adapter~\cite{chen2023vision}. ViT-Adapter, building upon the plain ViT~\cite{dosovitskiy2020image}, achieves precise image recognition by introducing image-specific inductive biases, which are also applied in many other areas~\cite{li2024algal, feng2024novel}. To perform effective TSD, we designed a traffic sign-specific output module based on ViT-Adapter. Besides, we constructed an extraction module to extract real images of traffic signs. The traffic sign-specific output module and extraction module compose the proposed TSD network.

\subsubsection{Traffic Sign-specific Output Module}\label{sub3.1.1}
\par We first take the original road image $I_o$ as an input into ViT-Adapter. The vanilla ViT-Adapter can generate segmentation images with various object category labels for the original input image. In our context, since we only need to detect traffic signs, it will interfere with the detection of traffic signs if segmentation images are generated for all object categories. Therefore, we designed a specialized output module for traffic signs. Specifically, in the generated segmentation image $I_c$, each specific object category is encoded with a different color for recognition. Our specialized output module converts $I_c$ into a binary mask image $I_b$. The method effectively distinguishes regions containing traffic signs from the rest of the image. The designed traffic sign-specific output module provides a clear delineation of the regions containing traffic signs and simplifies subsequent processing steps. 
The binary mask image $I_b$ effectively separates traffic signs from the background and other objects in $I_c$, thereby enhancing the dedicated detection capability for traffic signs.
\subsubsection{Traffic Sign Extraction Module}\label{sub3.1.2}
\par Next, we design an extraction module for realizing the extraction of real traffic signs. The extraction module consists of two parts. In the first part, we use the contour detection algorithm~\cite{suzuki1985topological} to accurately delineate and extract the coordinates of the boundaries of the traffic signs in the obtained binary mask image $I_b$. Specifically, given the binary mask image $I_b$ as a pixel grid in which each pixel is either foreground (representing the traffic sign) or background (representing the surroundings), the contour detection algorithm traces the boundaries of these components to efficiently outline the shape of the detected traffic sign $I_d$. $I_d$ includes the coordinate positions of traffic signs. In the second part, we use the original road image $I_o$ and the sign coordinates $I_d$ generated by the specific output mode of the ViT-Adapter to extract the image $I_r$ that contains only real traffic signs. Since the resolutions of $I_o$ and $I_d$ are the same, the image $I_r$ containing only real traffic signs is obtained by scanning the region of pixels containing white pixels in $I_d$ and then mapping the position of this region to the same region in $I_o$.
Finally, we extract the traffic sign image $I_t$ from $I_r$ using the coordinates of the traffic sign in the detected traffic sign $I_d$ calculated by the contour detection algorithm. $I_t$ is the final-detected traffic sign image of our TSD network. Note that although $I_t$ can also be obtained directly from the original road image $I_o$ through the coordinates obtained in $I_d$, the extracted traffic sign image contains the unnecessary background of the original image. In contrast, our extraction module avoids the potential interference caused by the background by extracting the background to black for accurate TSD.

\subsection{Cross-domain Few-shot In-context Learning for TSR}\label{sub3.2}
\par After implementing TSD and extracting the traffic sign image $I_t$, we perform cross-domain few-shot in-context learning TSR with MLLM. Typically, MLLMs take an image $I \in \mathbb{R}^{H \times W \times 3}$ and a text query $T_i = [t_i^1, \ldots, t_i^{l_i}]$ with length $l_i$, and generate a sequence of textual output $T_o = [t_o^1, \ldots, t_o^{l_o}]$ with length $l_o$ as follows:
\begin{equation}
T_o = \mathrm{MLLM}(I, T_i). 
\end{equation}
Since TSR is a fine-grained recognition task, different classes of traffic signs show similar characteristics. If only simply input the extracted traffic signs $I_t$ into the MLLM for recognition, although the MLLM can recognize the features of the input image, it is difficult for the MLLM to accurately identify the specific classes of traffic signs. To overcome this limitation and enhance fine-grained classification, we introduce a novel strategy. We generate textual descriptions of each traffic sign category by introducing template traffic signs and use the generated textual descriptions to assist MLLM in the fine-grained classification of traffic signs.
\par Specifically, in the proposed method, we first input the template traffic sign image $I^c_{\text{template}}$ into the MLLM to generate textual descriptions $T^c_d$ for each class $I^c_{\text{template}}=[I^1_{\text{template}}, ..., I^c_{\text{template}}]$ of traffic signs as follows:
\begin{equation}
T^c_d = \mathrm{MLLM}(I^c_{\text{template}}, T^c_i),
\end{equation}
where $T^c_i$ represents the corresponding input text query for the template traffic sign $I^c_{\text{template}}$. Thus, the textual descriptions of all classes $T_d$ are expressed as 
\begin{equation}
T_d = [T^1_d, ..., T^c_d]. 
\end{equation}

The template traffic signs are derived from the national standard traffic sign template database with standard shapes, colors, and features. The actual traffic sign images are diverse due to lighting conditions, angles, occlusions, etc., which are very different from the template traffic sign images, resulting in difficulties in cross-domain recognition. We instead utilize MLLM's powerful recognition of image features to reduce cross-domain differences by transforming images into textual descriptions. Meanwhile, since the proposed method requires only template traffic signs and MLLM generates descriptions only once for each template traffic sign class, the proposed few-shot method eliminates the dependence on large-scale training data with labels compared with previous methods. We input the extracted traffic sign image $I_t$, the generated text description of each template traffic sign class $T_d$, and the query text $T_i$ into the MLLM to obtain the following output $T_o$ of the proposed cross-domain few-shot in-context learning TSR method:
\begin{equation}
T_o = \mathrm{MLLM}(I_t, T_d, T_i).
\end{equation}
As we fully consider the three basic features of traffic signs i.e., shape, color, and composition when generating text descriptions, the generated text descriptions cover the key information of each traffic sign, thus assisting MLLM in fine-grained classification. Therefore, the fine-grained reasoning capability of MLLM for traffic signs is enhanced by the textual descriptions containing the main features generated by the proposed method.

\section{EXPERIMENTS}
\subsection{Experimental Settings}\label{sub4.1}
In this subsection, we explain the detailed experimental settings in this study. We conducted experiments on four different datasets, including two benchmark datasets: the German traffic sign recognition benchmark (GTSRB) dataset~\cite{stallkamp2012man} and the Belgium traffic sign dataset~\cite{mathias2013traffic}. Additionally, to fully evaluate the performance of the proposed method in real-world scenarios, we also performed experiments on two real-world road image datasets from Japan, namely the Sapporo urban road dataset and Yokohama urban road dataset. We used the MLLM of GPT4-o and GPT4-v for TSR, and we do not need to train any models for our method. However, due to the limited quota and the traffic limitations of the GPT's API~\footnote{https://platform.openai.com/ account/limits}, we followed the experimental setup method in~\cite{yang2023set} and randomly selected a subset of validation data from GTSRB and BTSD dataset for our study. Specifically, we selected 100 traffic sign images of 43 different types from GTSRB and 125 traffic sign images of 62 different types from BTSD. The number of categories in the subset was set to be the same as in the original dataset to fully validate the fine-grained recognition performance of our method. Since the traffic signs are already extracted in the two benchmark datasets, we utilized the proposed few-shot in-context learning method for fine-grained recognition. The Sapporo urban road dataset includes 68 images representing 18 different types of traffic signs, and the Yokohama urban road dataset includes 33 images with 13 types of traffic signs. For the two real-world datasets of Japan, we used the proposed TSD network to extract traffic signs from the original road images and performed TSR by the proposed MLLM-based method to evaluate the effectiveness of our method in real-world scenarios.

To evaluate the performance of the proposed TSR method, we adopt the evaluation metric Top-$k$ accuracy, which provides a comprehensive evaluation of the performance of the method. This metric can be defined as follows:

\begin{equation}
{\rm Top\mathchar`-}k = \frac{t_{k}}{\sum_{t} I_t}.\label{3}
\end{equation}
Here, \(t_{k}\) represents the number of correctly recognized input traffic images in the Top-$k$ recognition results. Considering the challenges of few-shot and zero-shot TSR since no training data is required, the Top-$k$ metric can achieve effective measurement of recognition performance. In addition, Top-$k$ can provide credible traffic sign candidates to assist drivers in judging road conditions in ADAS.
\begin{table*}[t]

\centering
\small
\setlength{\tabcolsep}{2pt}
\caption{Top-$k$ accuracy of different methods on four datasets.}
     \resizebox{0.95\textwidth}{!}{%
\begin{tabular}{l|ccc|ccc|ccc|ccc}
\toprule

\textbf{Method} &\multicolumn{3}{c|}{\textbf{GTSRB}} & \multicolumn{3}{c|}{\textbf{BTSD}} & \multicolumn{3}{c|}{\textbf{Sapporo urban road}} & \multicolumn{3}{c}{\textbf{Yokohama urban road}}\\
\midrule
& \textbf{Top-1} & \textbf{Top-5} & \textbf{Top-10} & \textbf{Top-1} & \textbf{Top-5} & \textbf{Top-10} & \textbf{Top-1} & \textbf{Top-5} & \textbf{Top-10} & \textbf{Top-1} & \textbf{Top-5} & \textbf{Top-10}\\
\midrule

Song~et~al.~\cite{yucong2021traffic} 
            &0.30  &0.56  &0.75 
            &0.72  &0.85  &0.88  
            &0.02  &0.30  &0.47 
            &0.03  &0.30  &0.39  
\\
Ren~et~al.~\cite{ren2009general} 
            &0.45  &0.74  &0.89
            &0.45  &0.71  &0.84 
            &0.13  &0.29  &0.31 
            &0.27  &0.46  &0.52  
\\

ResNet-50~\cite{he2016deep}   
            &0.63  &0.83  &0.83 
            &0.90  &0.95  &0.96  
            &0.34  &0.72  &0.79 
            &0.49  &0.68  &0.70  
\\

DenseNet-121~\cite{huang2017densely}  
            &0.60  &0.90  &0.97
            &0.66  &0.89  &0.93 
            &0.66  &0.80  &0.82 
            &0.67  &0.76  &0.79 
\\

EfficientNet-B0~\cite{tan2019efficientnet}
            &0.63  &0.94  &\textbf{1.00}
            &0.81  &0.94  &0.99  
            &0.37  &0.74  &0.79 
            &0.36  &0.64  &0.70  
\\

Mobilenet-V3~\cite{howard2019searching}
            &0.58  &0.95  &\textbf{1.00}
            &0.79  &0.98  &\textbf{1.00}
            &0.65  &0.82  &0.84 
            &0.64  &0.76  &0.79  
\\
MAE~\cite{he2022masked}
            &0.27  &0.51  &0.61 
            &0.50  &0.80  &0.86 
            &0.08  &0.47  &0.72
            &0.03  &0.39  &0.79
\\
CLIP~\cite{radford2021learning}
            &0.16  &0.42  &0.68 
            &0.22  &0.59  &0.71 
            &0.34  &0.57  &0.72
            &0.27  &0.64  &0.76 
 
\\

ViT-B~\cite{dosovitskiy2020image}
            &0.36  &0.68  &0.86
            &0.60  &0.77  &0.91
            &0.24  &0.72  &0.88  
            &0.24  &0.73  &0.91 
\\
ViT-L~\cite{dosovitskiy2020image}
            &0.60  &0.88  &0.98 
            &0.81  &0.96  &0.99 
            &0.44  &0.68  &0.81 
            &0.58  &0.67  &0.91 
\\

\textbf{Ours (GPT-4v)}
            &\textbf{0.88}  & \textbf{0.97}  & \textbf{1.00}
            &\textbf{0.91}  &  \textbf{0.98} & \textbf{1.00}  
            &\textbf{0.75}  & \textbf{0.84}  & \textbf{0.91}
            &\textbf{0.88}  &\textbf{0.94}   &\textbf{1.00} 
\\

\textbf{Ours (GPT-4o)}

            &\textbf{0.89}  & \textbf{0.98}  & 0.98
            &\textbf{0.91}  &  \textbf{0.99} & 0.99  
            &\textbf{0.82}  & \textbf{0.99}  & \textbf{1.00}
            &\textbf{0.93}  &\textbf{1.00}   &\textbf{1.00} 
\\
\bottomrule
\end{tabular}}
\label{tab1}
\end{table*}

\subsection{Experimental Results}\label{sub4.2}

\par Table~\ref{tab1} shows the Top-$k$ recognition results of the various methods. We perform a comprehensive evaluation on two benchmark datasets of traffic signs, GTSRB and BTSD, as well as two traffic sign datasets of Japan, to validate the effectiveness of the proposed method. As shown in Table~\ref{tab1}, the proposed method achieves promising results when compared to traditional methods as well as CNN-based TSR methods. The accuracy of Top-$k$ on four different datasets exceeds that of the comparative methods with substantial improvement, proving the effectiveness of the proposed method. In addition, experimental results demonstrate that the proposed MLLM-based method outperforms advanced transformer-based methods such as MAE, CLIP, ViT-B, and ViT-L. It is worth mentioning that the proposed method has a substantial lead in the recognition accuracy of Top-1 for all datasets compared to the CNN and transformer-based methods, which demonstrate the accurate recognition capability for traffic signs. Besides, our method shows promising performance in GPT-4o and GPT-4v, where GPT-4o outperforms GPT-4v. This demonstrates the potential of the proposed method to be easily extended to future versions of MLLM.
\par
\begin{figure*}[t]{
    \centering
    \includegraphics[width=1\textwidth]{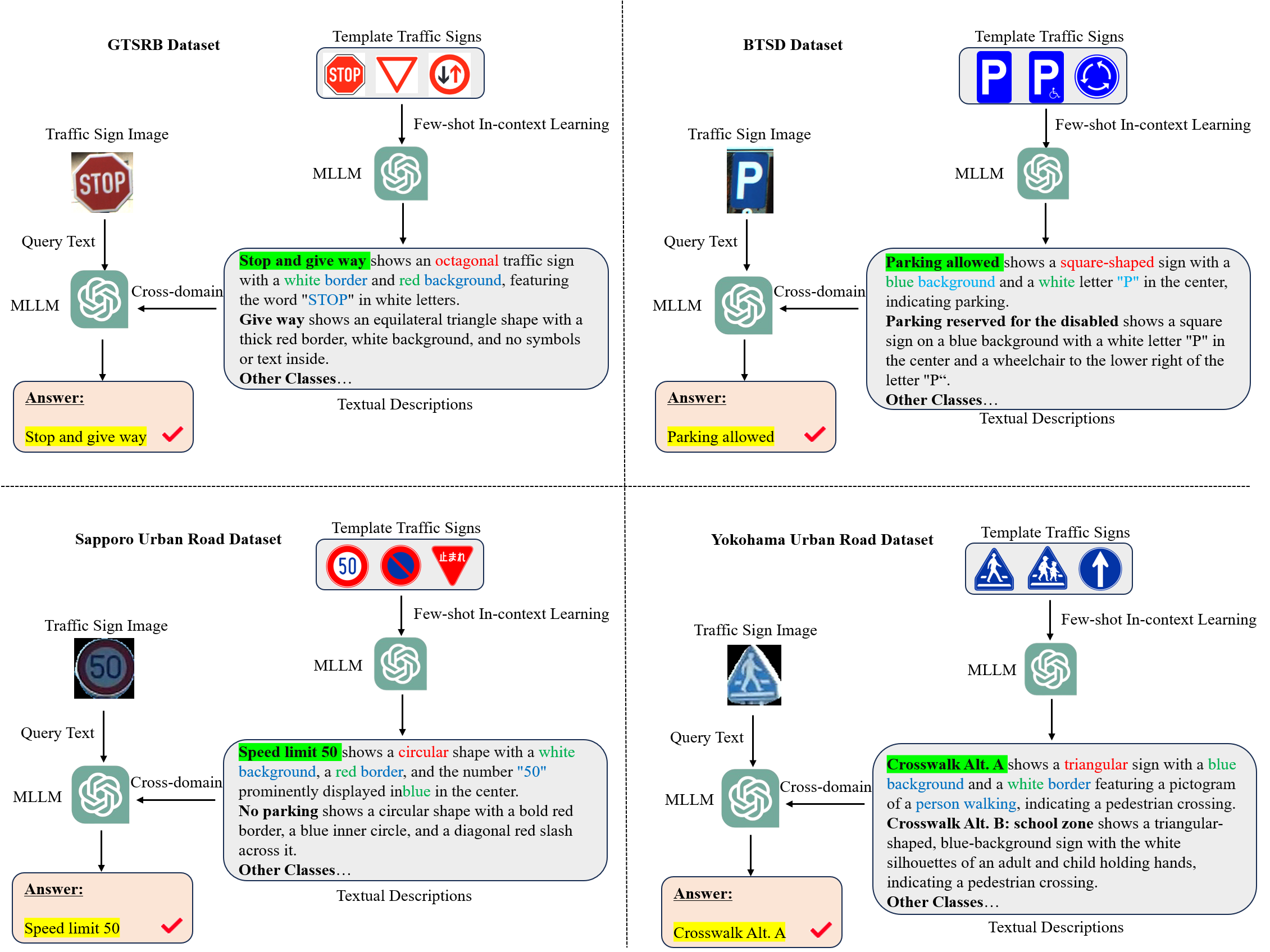}
    \caption{Top-1 recognition results of the proposed cross-domain few-shot in-context learning TSR method. We show samples for the benchmark dataset (GTSRB, BTSD) and the real-world dataset (Sapporo and Yokohama urban road dataset). In the Sapporo and Yokohama urban road datasets samples, we show the process from the TSD in original road images to the recognition of traffic signs by MLLM using the proposed method.} \label{fig2}}
\end{figure*}

\begin{figure*}[h]{
    \centering
    \includegraphics[width=1\textwidth]{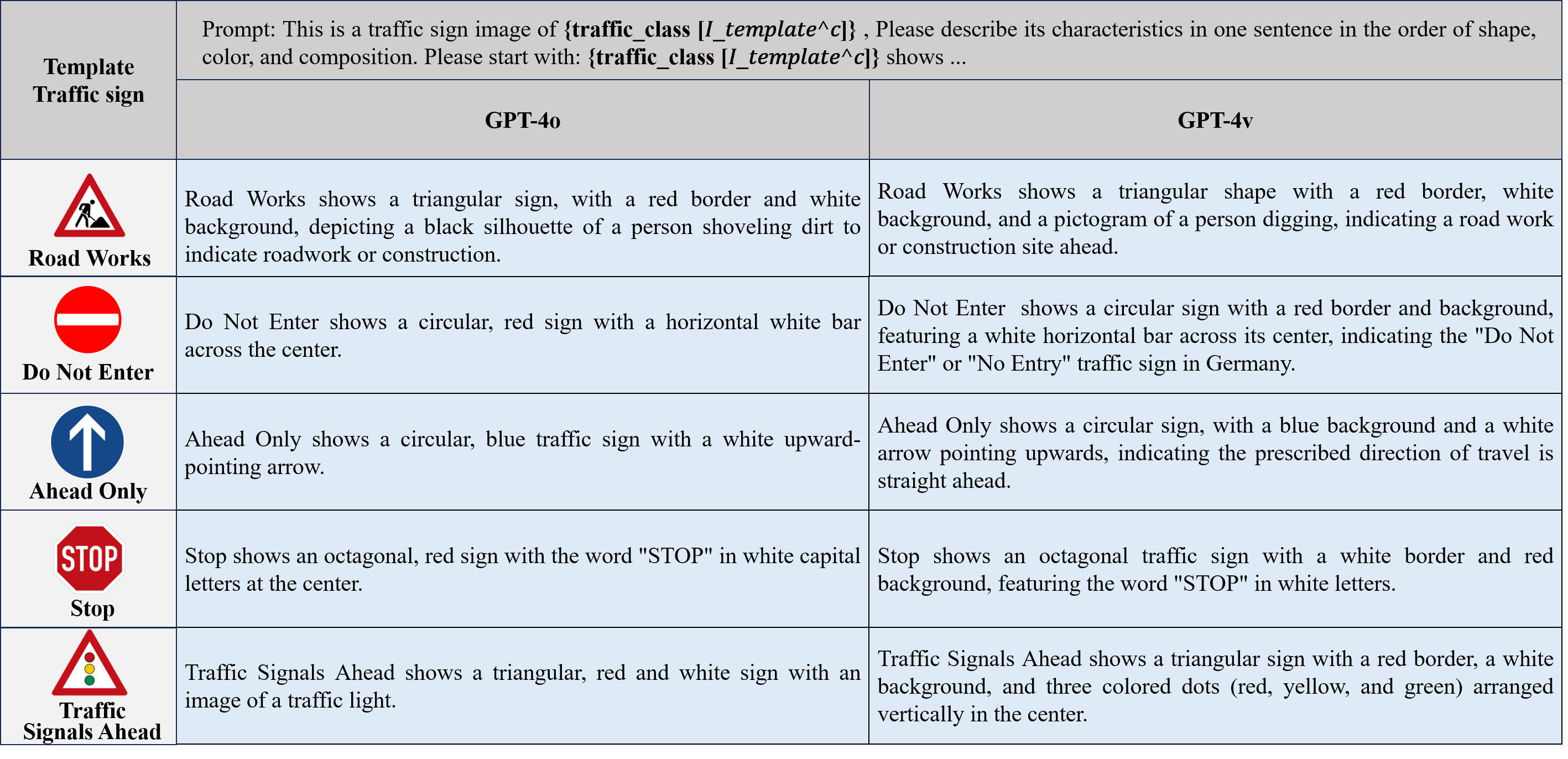}
    \caption{Examples of generated description texts for different MLLMs under the unified prompt (GTSRB dataset).} \label{fig5}}
\end{figure*}

\begin{figure*}[h]{
    \centering
    \includegraphics[width=1\textwidth]{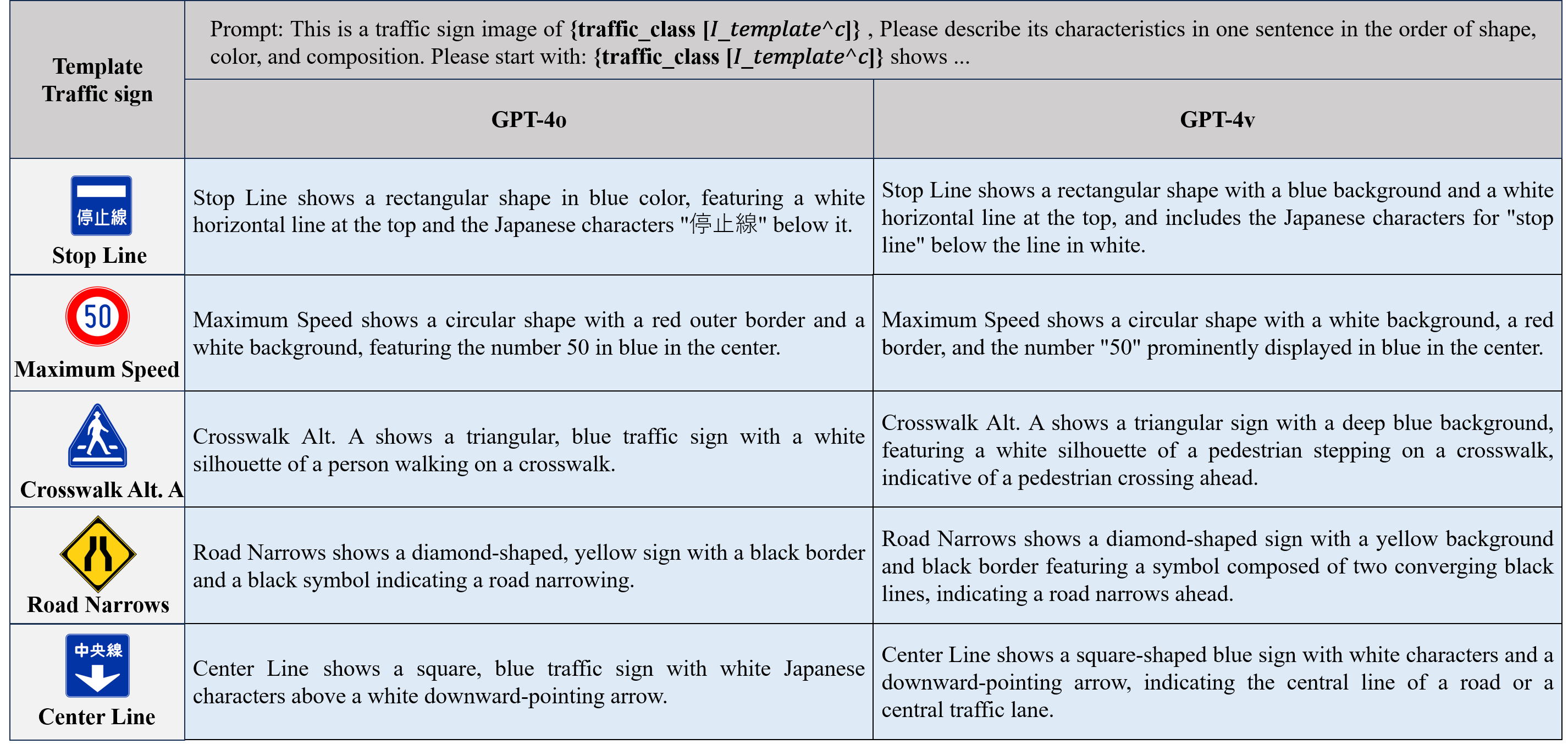}
    \caption{Examples of generated description texts for different MLLMs under the unified prompt (Sapporo urban road dataset).} \label{fig6}}
\end{figure*}

 Figure~\ref{fig2} shows examples of recognition results of the proposed method. We show two samples, one from a benchmark dataset and the other from a real-world dataset, to fully demonstrate the effectiveness of the proposed method. We show the original road images, input traffic signs, generated text descriptions, and cross-domain recognition results for MLLM. As shown in Fig.~\ref{fig2}, our method generates textual descriptions of traffic signs with simple and uniform prompts and template traffic sign images. Considering the characteristics of traffic signs, we design the input prompts for generating the text description so that the important features of the shape, color, and composition of traffic signs are fully considered. Then we use the generated text descriptions to enhance the fine-grained recognition capability of MLLM for traffic signs. Due to the differences between the template traffic sign and the actual traffic sign images, we reduce this cross-domain difference by transforming the images into textual descriptions using MLLM's powerful recognition of image features, thus achieving effective TSR. We further show the description texts generated by different MLLMs (GPT-4o and GPT-4v) under the unified prompt in Figures~\ref{fig5} and~\ref{fig6}. Compared to GPT-4v, the description texts generated by GPT-4o are generally shorter. This shows GPT-4o's better ability to refine the expression of traffic sign characteristics.

\begin{table*}[h]

\centering
\small
\setlength{\tabcolsep}{2pt}
\caption{Top-$k$ accuracy of directly input traffic sign image (baseline).}
     \resizebox{\textwidth}{!}{%
\begin{tabular}{l|ccc|ccc|ccc|ccc}
\toprule

\textbf{Method} &\multicolumn{3}{c|}{\textbf{GTSRB}} & \multicolumn{3}{c|}{\textbf{BTSD}} & \multicolumn{3}{c|}{\textbf{Sapporo urban road}} & \multicolumn{3}{c}{\textbf{Yokohama urban road}}\\
\midrule
& \textbf{Top-1} & \textbf{Top-5} & \textbf{Top-10} & \textbf{Top-1} & \textbf{Top-5} & \textbf{Top-10} & \textbf{Top-1} & \textbf{Top-5} & \textbf{Top-10} & \textbf{Top-1} & \textbf{Top-5} & \textbf{Top-10}\\
\midrule

Baseline (GPT-4v)
            &0.80  &0.86  &0.87
            &0.61  &0.70  &0.75 
            &0.28  &0.43  &0.54
            &0.46  &0.70  &0.76 
\\
\textbf{Ours (GPT-4v)}
            &\textbf{0.88}  & \textbf{0.97}  & \textbf{1.00}
            &\textbf{0.91}  &  \textbf{0.98} & \textbf{1.00}  
            &\textbf{0.75}  & \textbf{0.84}  & \textbf{0.91}
            &\textbf{0.88}  &\textbf{0.94}   &\textbf{1.00} 
\\
Baseline (GPT-4o)
            &0.88  &0.89  &0.89
            &0.79  &0.89  &0.89 
            &0.46  &0.66  &0.80
            &0.65  &0.88  &0.88 

\\
\textbf{Ours (GPT-4o)}
            &\textbf{0.89}  & \textbf{0.98}  & \textbf{0.98}
            &\textbf{0.91}  &  \textbf{0.99} & \textbf{0.99}  
            &\textbf{0.82}  & \textbf{0.99}  & \textbf{1.00}
            &\textbf{0.93}  &\textbf{1.00}   &\textbf{1.00} 
\\

\bottomrule
\end{tabular}}
\label{tab2}
\end{table*}

\begin{figure*}[ht]{
    \centering
    \includegraphics[width=1\textwidth]{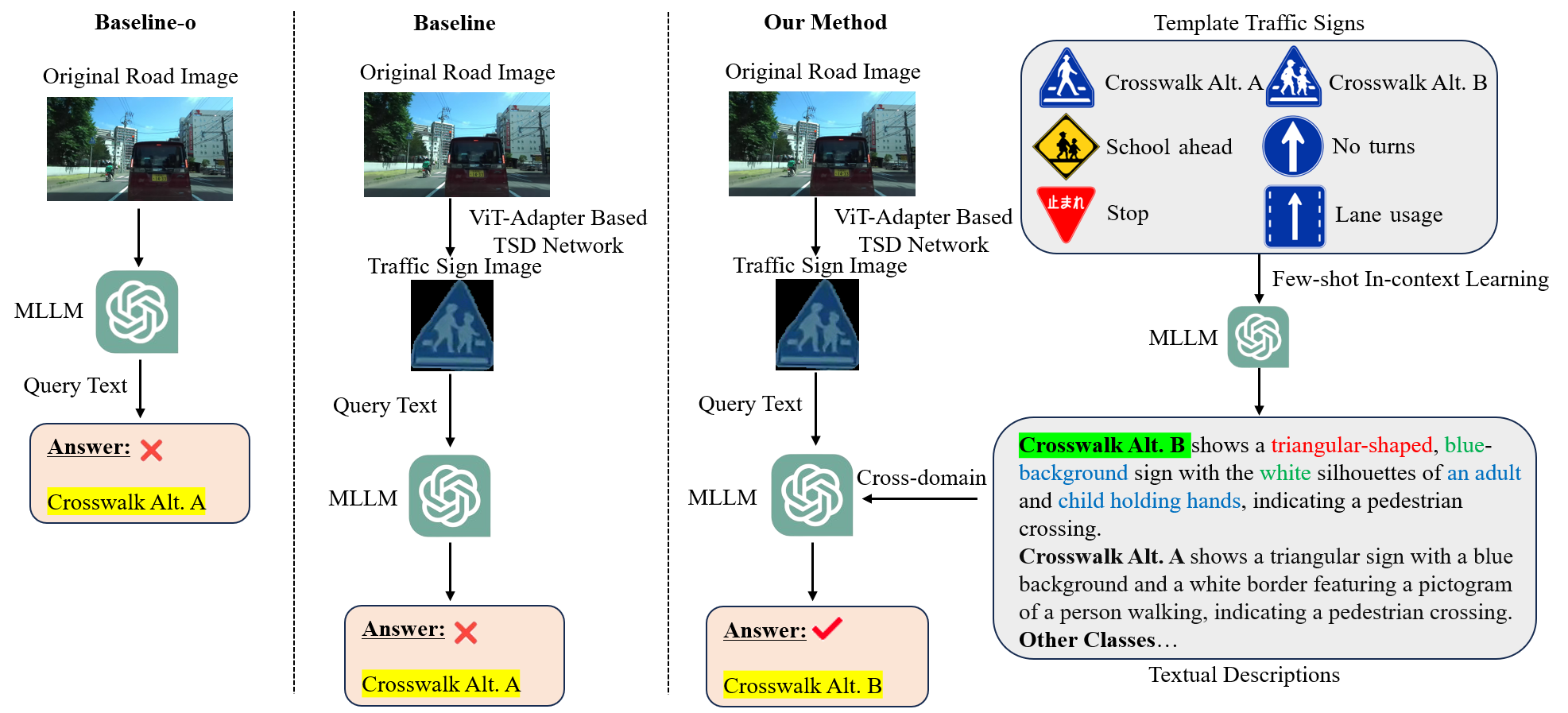}
    \caption{Top-1 recognition results of the baseline-o, baseline, and our method.} \label{fig3}}
\end{figure*}

\begin{figure}[t]{
    \centering
    \includegraphics[width=0.8\textwidth]{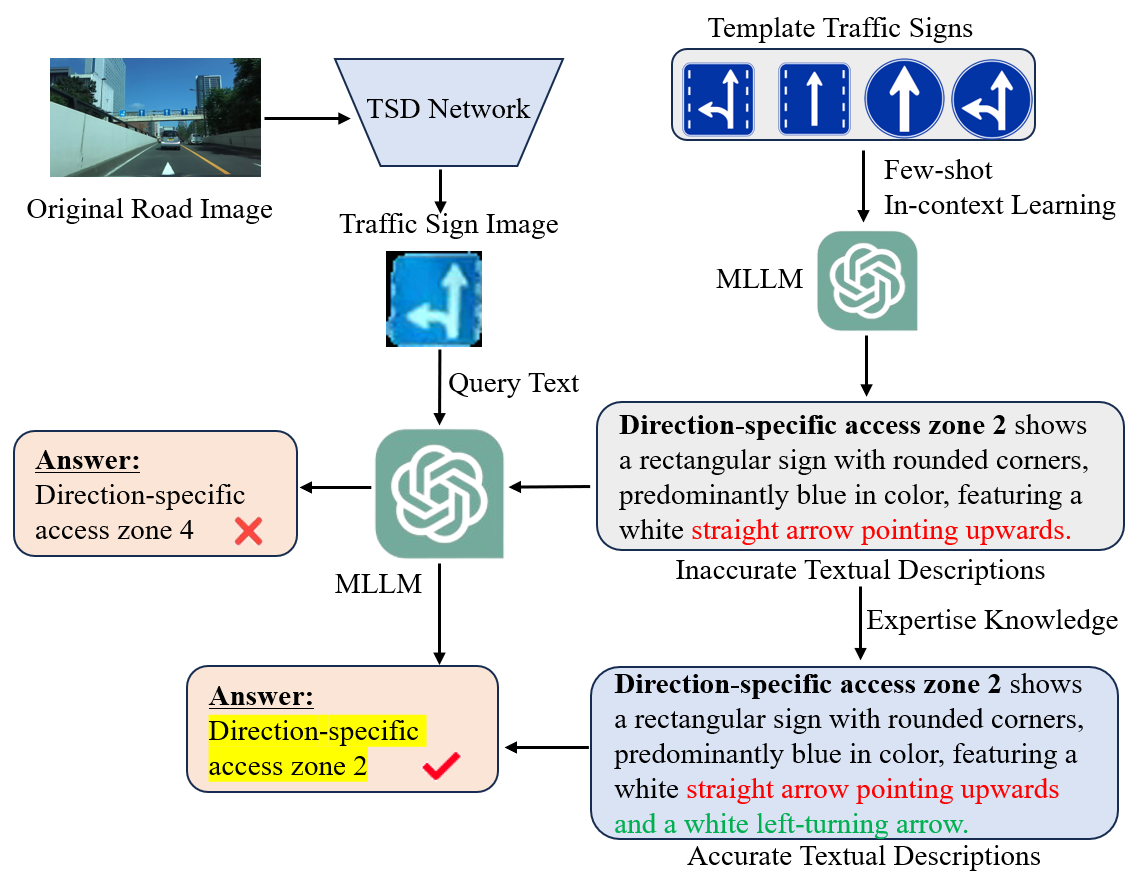}
    \caption{Example of incorrect recognition due to inaccurate generated description texts.} \label{fig4}}
\end{figure}
\subsection{Ablation Studies}\label{sub4.3}
 Furthermore, we validate the TSR results of inputting traffic sign image directly (baseline) as shown in Table~\ref{tab2}, and the Top-$k$ accuracy of the baseline is relatively low on all four datasets, indicating that it is difficult to perform accurate fine-grained recognition of traffic signs relying only on the MLLM. Our method on cross-domain template traffic signs provides key information about the shape, color, and composition of traffic signs, which enhances the fine-grained perception of traffic signs in MLLM for better TSR. 
 \par In addition to the MLLM-based baseline method of inputting only traffic sign images in Table~\ref{tab2}, there is another recognition method commonly used for other tasks, i.e., the original road images containing traffic signs are directly input into the MLLM for recognition, name as baseline-o. Since there is no relevant TSR study based on MLLM, the effectiveness of this method is not yet known. Therefore, to further demonstrate the effectiveness of the proposed method, we conducted an ablation study. We exemplify the recognition results of three MLLM-based methods (input original road image including traffic signs (baseline-o), input traffic sign image (baseline), and our method) on the same traffic sign. As shown in Fig.~\ref{fig3}, the original road image is directly input into MLLM for recognition in the baseline-o method. Although the original road image may contain the contextual information of the road, the traffic signs are small and the fine-grained traffic sign features are very similar, such as Crosswalk Alt. A and Crosswalk Alt. B in the sample. Therefore, the direct use of the MLLM for TSR does not perform well. Besides, although the baseline method extracts the traffic signs using our TSD network, it performs poorly because it is difficult for MLLM to recognize the specific types of traffic signs. In contrast, the proposed method generates descriptions of different types of traffic signs by introducing template traffic signs, which enhances the fine-grained recognition ability of traffic signs by MLLM.
\par In addition, we show an example of incorrect recognition due to the generation of inaccurate text descriptions. As shown in Fig.~\ref{fig4}, the traffic sign Direction-specific access zone 2 has straight and left turn arrows, while the generated description texts contain only straight arrows. Since key information about the leftward arrow of Direction-specific access zone 2 is missing, it is incorrectly recognized as Direction-specific access zone 4 by MLLM. The results show the importance of accurate description texts for the fine-grained recognition of MLLM. Accurate recognition was achieved by introducing the expertise knowledge to correct the description texts.

\section{Conclusion}
\par We have proposed a novel cross-domain few-shot in-context learning method based on MLLM for enhancing TSR in this paper. We proposed a new TSD network and generated textual descriptions by introducing template traffic signs to enhance the fine-grained recognition of traffic signs by MLLM. Experimental results conducted on two benchmark datasets and two real-world datasets demonstrate the effectiveness of the proposed method. This paper is the first exploration of enhancing MLLM's ability to recognize traffic signs at a fine-grained level, and we hope to inspire TSR-related research based on MLLM in the community.

\section*{Acknowledgement}
%
Some data in this study were provided by Japan Radio Co., Ltd. This study was supported in part by JSPS KAKENHI Grant Numbers JP23K21676, JP23K11211, JP23K11141, and JST SPRING, Grant Number JPMJSP2119.
\bibliographystyle{plain}
\bibliography{neurips_2024}

\begin{thebibliography}{10}

\bibitem{akatsuka1987road}
Hidehiko Akatsuka and Shinichiro Imai.
\newblock Road signposts recognition system.
\newblock {\em SAE Technical Paper}, pages 1--12, 1987.

\bibitem{bubeck2023sparks}
S{\'e}bastien Bubeck, Varun Chandrasekaran, Ronen Eldan, Johannes Gehrke, Eric Horvitz, Ece Kamar, Peter Lee, Yin~Tat Lee, Yuanzhi Li, Scott Lundberg, et~al.
\newblock Sparks of artificial general intelligence: Early experiments with gpt-4.
\newblock {\em arXiv}, 2023.

\bibitem{chang2023survey}
Yupeng Chang, Xu~Wang, Jindong Wang, Yuan Wu, Kaijie Zhu, Hao Chen, Linyi Yang, Xiaoyuan Yi, Cunxiang Wang, Yidong Wang, et~al.
\newblock A survey on evaluation of large language models.
\newblock {\em arXiv}, 2023.

\bibitem{chen2005tutorial}
Pai-Hsuen Chen, Chih-Jen Lin, and Bernhard Sch{\"o}lkopf.
\newblock A tutorial on $\nu$-support vector machines.
\newblock {\em Applied Stochastic Models in Business and Industry}, 21(2):111--136, 2005.

\bibitem{chen2023vision}
Zhe Chen, Yuchen Duan, Wenhai Wang, Junjun He, Tong Lu, Jifeng Dai, and Yu~Qiao.
\newblock Vision transformer adapter for dense predictions.
\newblock In {\em Proc. ICLR}, 2023.

\bibitem{cui2024drive}
Can Cui, Yunsheng Ma, Xu~Cao, Wenqian Ye, and Ziran Wang.
\newblock Drive as you speak: Enabling human-like interaction with large language models in autonomous vehicles.
\newblock In {\em Proc. CVPR}, pages 902--909, 2024.

\bibitem{dilek2023computer}
Esma Dilek and Murat Dener.
\newblock Computer vision applications in intelligent transportation systems: a survey.
\newblock {\em Sensors}, 23(6):2938, 2023.

\bibitem{dosovitskiy2020image}
Alexey Dosovitskiy, Lucas Beyer, Alexander Kolesnikov, Dirk Weissenborn, Xiaohua Zhai, Thomas Unterthiner, Mostafa Dehghani, Matthias Minderer, Georg Heigold, Sylvain Gelly, et~al.
\newblock An image is worth 16x16 words: Transformers for image recognition at scale.
\newblock In {\em Proc. ICLR}, 2020.

\bibitem{feng2024novel}
Yuhu Feng, Jiahuan Zhang, Guang Li, Ren Togo, Keisuke Maeda, Takahiro Ogawa, and Miki Haseyama.
\newblock A novel frame-selection metric for video inpainting to enhance urban feature extraction.
\newblock {\em Sensors}, 24(10):3035, 2024.

\bibitem{economic1968convention}
Economic~Commission for Europe-Inland Tansport~Committee et~al.
\newblock Convention on road signs and signals.
\newblock {\em United Nations Treaty Series}, 1091:3, 1968.

\bibitem{gan2023zero}
Yaozong Gan, Guang Li, Ren Togo, Keisuke Maeda, Takahiro Ogawa, and Miki Haseyama.
\newblock Zero-shot traffic sign recognition based on midlevel feature matching.
\newblock {\em Sensors}, 23(23):9607, 2023.

\bibitem{hammoudi2022road}
Abdulla Hammoudi, George Karani, and John Littlewood.
\newblock Road traffic accidents among drivers in abu dhabi, united arab emirates.
\newblock {\em Journal of Traffic and Logistics Engineering}, pages 1--6, 2022.

\bibitem{he2022masked}
Kaiming He, Xinlei Chen, Saining Xie, Yanghao Li, Piotr Doll{\'a}r, and Ross Girshick.
\newblock Masked autoencoders are scalable vision learners.
\newblock In {\em Proc. CVPR}, pages 16000--16009, 2022.

\bibitem{he2016deep}
Kaiming He, Xiangyu Zhang, Shaoqing Ren, and Jian Sun.
\newblock Deep residual learning for image recognition.
\newblock In {\em Proc. CVPR}, pages 770--778, 2016.

\bibitem{howard2019searching}
Andrew Howard, Mark Sandler, Grace Chu, Liang-Chieh Chen, Bo~Chen, Mingxing Tan, Weijun Wang, Yukun Zhu, Ruoming Pang, Vijay Vasudevan, et~al.
\newblock Searching for mobilenetv3.
\newblock In {\em Proc. CVPR}, pages 1314--1324, 2019.

\bibitem{hua2010traffic}
Xiaoguang HUa, Xinyan ZHUa, Deren LIa, and Hui LI.
\newblock Traffic sign recognition using scale invariant feature transform and svm.
\newblock In {\em Proc. ISPRS}, pages 15--19, 2010.

\bibitem{huang2017densely}
Gao Huang, Zhuang Liu, Laurens Van Der~Maaten, and Kilian~Q Weinberger.
\newblock Densely connected convolutional networks.
\newblock In {\em Proc. CVPR}, pages 4700--4708, 2017.

\bibitem{kerim2021recognition}
Abdulrahman Kerim and Mehmet~{\"O}nder Efe.
\newblock Recognition of traffic signs with artificial neural networks: A novel dataset and algorithm.
\newblock In {\em Proc. ICAIIC}, pages 171--176, 2021.

\bibitem{kus2008traffic}
Merve~Can Kus, Muhittin Gokmen, and Sima Etaner-Uyar.
\newblock Traffic sign recognition using scale invariant feature transform and color classification.
\newblock In {\em Proc. ISCIS}, pages 1--6, 2008.

\bibitem{li2024algal}
Guang Li, Ren Togo, Keisuke Maeda, Akinori Sako, Isao Yamauchi, Tetsuya Hayakawa, Shigeyuki Nakamae, Takahiro Ogawa, and Miki Haseyama.
\newblock Algal bed region segmentation based on a vit adapter using aerial images for estimating co2 absorption capacity.
\newblock {\em Remote Sensing}, 16(10):1742, 2024.

\bibitem{li2022finely}
Wei Li, Haiyu Song, and Pengjie Wang.
\newblock Finely crafted features for traffic sign recognition.
\newblock {\em International Journal of Circuits, Systems and Signal Processing}, 16:159--170, 2022.

\bibitem{lowe2004distinctive}
David~G Lowe.
\newblock Distinctive image features from scale-invariant keypoints.
\newblock {\em International Journal of Computer Vision}, 60:91--110, 2004.

\bibitem{luo2017traffic}
Hengliang Luo, Yi~Yang, Bei Tong, Fuchao Wu, and Bin Fan.
\newblock Traffic sign recognition using a multi-task convolutional neural network.
\newblock {\em IEEE Transactions on Intelligent Transportation Systems}, 19(4):1100--1111, 2017.

\bibitem{mathias2013traffic}
Markus Mathias, Radu Timofte, Rodrigo Benenson, and Luc Van~Gool.
\newblock Traffic sign recognition—how far are we from the solution?
\newblock In {\em Proc. IJCNN}, pages 1--8, 2013.

\bibitem{radford2021learning}
Alec Radford, Jong~Wook Kim, Chris Hallacy, Aditya Ramesh, Gabriel Goh, Sandhini Agarwal, Girish Sastry, Amanda Askell, Pamela Mishkin, Jack Clark, et~al.
\newblock Learning transferable visual models from natural language supervision.
\newblock In {\em Proc. ICML}, pages 8748--8763, 2021.

\bibitem{ren2009general}
FeiXiang Ren, Jinsheng Huang, Ruyi Jiang, and Reinhard Klette.
\newblock General traffic sign recognition by feature matching.
\newblock In {\em Proc. IVCNZ}, pages 409--414, 2009.

\bibitem{romdhane2016improved}
Nadra~Ben Romdhane, Hazar Mliki, and Mohamed Hammami.
\newblock An improved traffic signs recognition and tracking method for driver assistance system.
\newblock In {\em Proc. ICIS}, pages 1--6, 2016.

\bibitem{saadna2017overview}
Yassmina Saadna and Ali Behloul.
\newblock An overview of traffic sign detection and classification methods.
\newblock {\em International Journal of Multimedia Information Retrieval}, 6:193--210, 2017.

\bibitem{seraj2021implications}
Mudasser Seraj, Andres Rosales-Castellanos, Amr Shalkamy, Karim El-Basyouny, and Tony~Z Qiu.
\newblock The implications of weather and reflectivity variations on automatic traffic sign recognition performance.
\newblock {\em Journal of Advanced Transportation}, pages 1--15, 2021.

\bibitem{stallkamp2012man}
Johannes Stallkamp, Marc Schlipsing, Jan Salmen, and Christian Igel.
\newblock Man vs. computer: Benchmarking machine learning algorithms for traffic sign recognition.
\newblock {\em Neural Networks}, 32:323--332, 2012.

\bibitem{supriyanto2016unsupervised}
Catur Supriyanto, Ardytha Luthfiarta, and Junta Zeniarja.
\newblock An unsupervised approach for traffic sign recognition based on bag-of-visual-words.
\newblock In {\em Proc. ICITEE}, pages 1--4, 2016.

\bibitem{suzuki1985topological}
Satoshi Suzuki et~al.
\newblock Topological structural analysis of digitized binary images by border following.
\newblock {\em Computer Vision, Graphics, and Image Processing}, 30(1):32--46, 1985.

\bibitem{tan2019efficientnet}
Mingxing Tan and Quoc Le.
\newblock Efficientnet: Rethinking model scaling for convolutional neural networks.
\newblock In {\em Proc. ICML}, pages 6105--6114, 2019.

\bibitem{yang2023set}
Jianwei Yang, Hao Zhang, Feng Li, Xueyan Zou, Chunyuan Li, and Jianfeng Gao.
\newblock Set-of-mark prompting unleashes extraordinary visual grounding in gpt-4v.
\newblock {\em arXiv}, 2023.

\bibitem{yang2023dawn}
Zhengyuan Yang, Linjie Li, Kevin Lin, Jianfeng Wang, Chung-Ching Lin, Zicheng Liu, and Lijuan Wang.
\newblock The dawn of lmms: Preliminary explorations with gpt-4v (ision).
\newblock {\em arXiv}, 2023.

\bibitem{1520577215790999680}
Gan Yaozong, Li~Guang, Togo Ren, Maeda Keisuke, Ogawa Takahiro, and Haseyama Miki.
\newblock A note on traffic sign recognition based on vision transformer adapter using visual feature matching.
\newblock {\em ITE technical report}, 47(6):1--4, 2023.

\bibitem{you2023ferret}
Haoxuan You, Haotian Zhang, Zhe Gan, Xianzhi Du, Bowen Zhang, Zirui Wang, Liangliang Cao, Shih-Fu Chang, and Yinfei Yan.
\newblock Ferret: Refer and ground anything anywhere at any granularity.
\newblock {\em arXiv}, 2023.

\bibitem{yucong2021traffic}
Song Yucong and Guo Shuqing.
\newblock Traffic sign recognition based on hog feature extraction.
\newblock {\em Journal of Measurements in Engineering}, 9(3):142--155, 2021.

\bibitem{zhang2023llama}
Renrui Zhang, Jiaming Han, Aojun Zhou, Xiangfei Hu, Shilin Yan, Pan Lu, Hongsheng Li, Peng Gao, and Yu~Qiao.
\newblock Llama-adapter: Efficient fine-tuning of language models with zero-init attention.
\newblock {\em arXiv}, 2023.

\bibitem{zhao2023survey}
Wayne~Xin Zhao, Kun Zhou, Junyi Li, Tianyi Tang, Xiaolei Wang, Yupeng Hou, Yingqian Min, Beichen Zhang, Junjie Zhang, Zican Dong, et~al.
\newblock A survey of large language models.
\newblock {\em arXiv}, 2023.

\bibitem{zheng2022evaluation}
Yuping Zheng and Weiwei Jiang.
\newblock Evaluation of vision transformers for traffic sign classification.
\newblock {\em Wireless Communications and Mobile Computing}, 2022, 2022.

\bibitem{zhu2016traffic}
Zhe Zhu, Dun Liang, Songhai Zhang, Xiaolei Huang, Baoli Li, and Shimin Hu.
\newblock Traffic-sign detection and classification in the wild.
\newblock In {\em Proc. CVPR}, pages 2110--2118, 2016.

\end{thebibliography}
\newpage
\appendix

\end{document}